\documentclass[a4paper,twoside,french]{article}
\usepackage{orasis}
\usepackage[utf8]{inputenc}
\usepackage[T1]{fontenc}
\usepackage{times}
\usepackage{authblk}
\usepackage[fleqn]{amsmath}
\usepackage[table]{xcolor}
\usepackage{graphicx}
\graphicspath{ {./graphics/} }

\begin{document}
\date{}
\title{\Large\bf Adam SLAM - the last mile of camera calibration with 3DGS}
\author[1,2]{Matthieu GENDRIN}
\author[1]{Stéphane PATEUX}
\author[2]{Xiaoran JIANG}
\author[1]{Théo LADUNE}
\author[2]{Luce MORIN}
\affil[1]{Orange Innovation, France, {\fontfamily{qcr}\selectfont\small firstname.lastname@orange.com}}
\affil[2]{Univ Rennes, INSA Rennes, CNRS, IETR (UMR 6164), France, {\fontfamily{qcr}\selectfont\small firstname.lastname@insa-rennes.fr}}
\maketitle
\subsection*{Abstract}
{\em
The quality of the camera calibration is of major importance for evaluating progresses in novel view synthesis,
as a 1-pixel error on the calibration has a significant impact on the reconstruction quality.
While there is no ground truth for real scenes, the quality of the calibration is assessed by the quality of the novel view synthesis.
This paper proposes to use a 3DGS \cite{kerbl20233d} model to fine tune calibration by backpropagation of novel view color loss with
respect to the cameras parameters. The new calibration alone brings an average improvement of 0.4 dB PSNR on the dataset used as reference by 3DGS \cite{kerbl20233d}.
The fine tuning may be long and its suitability depends on the criticity of training time, but for calibration
of reference scenes, such as \cite{barron2022mip}, the stake of novel view quality is the most important.
}
\subsection*{Keywords}
Example, model, template.

\section{Introduction}
3d gaussian splatting \cite{kerbl20233d} has been a major breakthrough in the domain of 3D novel view synthesis, reaching state-of-the-art quality with an
efficient and realtime rendering performance. The research community has been florishing since, offering continuous but modest improvements.
For instance, MCMC \cite{kheradmand20243d} improvements range from 0.4 dB and 0.7 dB of PSNR (Peak Signal to Noise Ratio, a reference
quality metric in novel view synthesis).

In such a context, the quality of the camera calibration is of major importance, as a 1-pixel error on the calibration has a significant
impact on the reconstruction quality, sometimes up to 1 dB PSNR. The calibration is a field of research in itself, which does not focus on quality but seeks 
the best balance between precision, robustness and computational efficiency. The state of the art is held by feature based algorithms,
with COLMAP \cite{schonberger2018robust} as a reference, and many proposals to improve the features tracks used in that algorithm (see e.g. \cite{lindenberger2021pixel}).
Such solutions give a calibration suitable for novel view synthesis algorithms to converge. However, the calibration might introduce errors
in some circumstances, e.g. low-textured scenes, or with uneven features distribution. 
The hard point about evaluating the quality of calibration is that there is no ground truth for real scenes. As such, the quality of the calibration is assessed
by the quality of the novel view synthesis. Some publications leverage amortized learned neural networks, although
mainly to improve processing performance. DUSt3R \cite{wang2024dust3r} for example claims to avoid the cumbersome estimate of camera parameters,
but does not claim better final quality.

Taking this into consideration, previous works propose to train the pose of the cameras based on the loss of the novel view synthesis itself, like 
gsplatloc \cite{zeller2024gsplatloc} which generates a depthmap from a 3DGS model and compares it with a ground truth, or gsplat \cite{ye2024gsplat}
which backpropagates on the pose of the camera during the 3DGS training. Similar to \cite{ye2024gsplat}, we propose to train a 3DGS model from
a pre-existing calibration, and fine-tune this calibration based on the color loss of the novel view synthesis. Unlike gsplat,
we focus some iterations of the training on calibration improvement, replace L1 loss by a L2 loss in these iterations which performs better than L1,
and train the focal length of the cameras alongside the pose. We also propose a reparameterization of the camera extrinsics and intrinsics
for a more efficient training. These changes improve the final quality of the calibration, reaching an average 0.4 dB PSNR improvement
on the reference dataset of 3DGS \cite{kerbl20233d}.

\section{Method} \label{method}

This paper proposes to use backpropagation to train camera parameters from the color loss.
Section \ref{backprop} explains how backpropagation is simplified for implementation ease.
The choice of the color loss is discussed in section \ref{L2}.
Then the training schedule between model and camera fine-tuning is the subject of \ref{schedule}.
Finally, section \ref{abc} proposes a reparameterization of the camera model for a more performant training with Adam optimizer.

\subsection{Backpropagation} \label{backprop}
The whole rendering is kept identical to 3DGS \cite{kerbl20233d}, except that the camera parameters are trained based on the color loss.
We note the camera parameters $x_c,y_c,z_c,q,\phi_x,\phi_y$ where $x_c,y_c,z_c$ and $q$ are respectively the translation and rotation (as a quaternion) of the
world-to-camera pose, and $\phi_x,\phi_y$ are the fields of view along $u,v$ axis.
Noting $p_j$ one parameter of camera $j$ listed above, the derivatives of the color loss $\mathcal{L}$ with respect to $p$ are:
\begin{align}
	\label{eq:1}
	\frac{\partial \mathcal{L}}{\partial p_j} &= \sum_{i} \frac{\partial \mathcal{L}}{\partial uv_i^j} \frac{\partial uv_i^j}{\partial p_j} + \frac{\partial \mathcal{L}}{\partial cov2d_i^j} \frac{\partial cov2d_i^j}{\partial p_j}
\end{align}
With $uv_i^j$ and $cov2d_i^j$ the projections on image plan of the camera $j$ of respectively the center and the covariance matrix of the gaussian $i$.

We neglect the impact of $p_j$ on $cov2d_i^j$ in \eqref{eq:1}, leading to:
\begin{align}
	\label{eq:2}
	\frac{\partial \mathcal{L}}{\partial p_j} &= \sum_{i} \frac{\partial \mathcal{L}}{\partial uv_i^j} \frac{\partial uv_i^j}{\partial p_j}
\end{align}

This approximation is valid if the covariance of gaussian $i$ is small compared to its distance to the camera $j$,
which is the case for most gaussians in the model. Note that this approximation is already done by \cite{kerbl20233d}
when they approximate pinhole projection of the gaussian by an orthogonal projection.
In practice, Colmap's lack of precision usually happens on the $z_c$ position versus the $\phi_x,\phi_y$.
Changing these parameters will have no impact on the transformation of the covariance from absolute referential
to the relative referential, and small impact on the projection to the image plan.

\subsection{Choice of the loss} \label{L2}
While the choice of L1 loss (MAE: Mean Absolute Error)
achieves competitive results in training gaussian model, we
found empirically that L2 error (MSE: Mean Square Error)
is more suited to train the camera parameters. The choice
of the appropriate loss is a trade-off between the convergence
performance of L2 and the robustness to outliers
offered by L1. On a total of a few tens to few hundreds
of cameras, a mis-calibrated camera may not be negligible
for the training of the gaussians it sees. This explains
the choice of L1 for 3DGS \cite{kerbl20233d}. On the contrary, fine-tuning
the calibration of one camera based on the millions of gaussians
that compose the 3D model is less subject to outliers,
thus giving an advantage to the L2 loss. In practice, our
experiments show that L1 can lead the system to diverge,
some scenes ending up with trained cameras parameters
worse than the input ones. \\
Using different losses for model training and camera calibration
implies to dedicate some iterations to camera parameters
training. The results are more precise and reliable,
at the cost of an extended training duration. Though processing
performance is not our main focus, we propose a
training schedule in the next section to mitigate this problem.

\subsection{Training schedule} \label{schedule}
The 3DGS model training uses the cameras calibration as an input, and vice versa. Concurrent training of interdependent
parameters are common in machine learning, and gsplat \cite{ye2024gsplat} trains both the camera parameters and the 3DGS
model in parallel.
We propose a different schedule, by phase, with an early stopping mechanism to save compute time on cameras which do not 
benefit from fine-tuning. Specifically, we proceed in a loop with $M$ iterations of 3DGS model training, followed by the training of all the cameras parameters, one camera
after another. One camera is trained for at least 100 steps and maximum 1000 steps, a progress score being evaluated at every step and
the training of the camera is interrupted when the progress score reaches a minimum threshold $t$.

The progress score $s$ is an exponential moving average (EMA) of the PSNR progress:
\begin{align}
	s_{n+1} &= s_n (1 - \beta) + \beta (PSNR_{n+1} - PSNR_n) 
\end{align}
In our experiments, we initialize the score with $s_0 = 0$ and use $\beta = \frac{1}{50}$. 
This ensures that after the first 100 steps the EMA is a good approximation of the average progress.
Letting $M = 3000$ model training steps between cameras
fine-tuning is the natural pace of 3DGS \cite{kerbl20233d} training, as it
is the default interval between opacity reset. We experimentally found that the threshold $t = 0.0002$ was a good
trade-off between training duration and final quality.\\
This schedule enables us to save training iterations on cameras whose training does not improve the PSNR.
In our experiments, this leads to 20\% savings in cameras training time.

\begin{figure}
  \includegraphics[width=\linewidth]{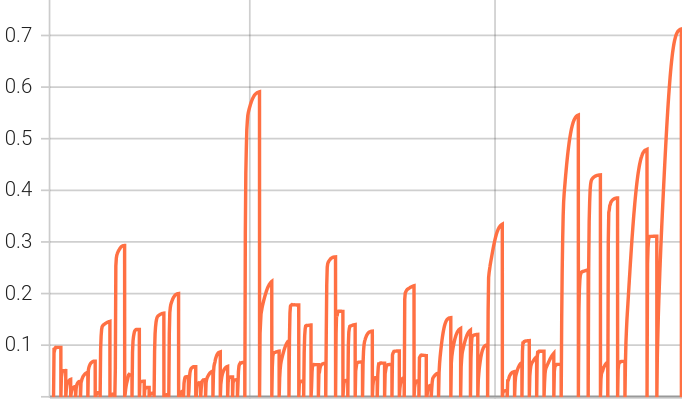}
  \caption{Tracking of the PSNR improvement during the first phase of cameras fine-tuning. Each increasing subgraph concerns one camera.
  One can see that the duration of training depends on the improvement pace. Y axis is the PSNR improvement of the current camera, x axis
  is the iteration.}
\end{figure}

\subsection{Reparameterization} \label{abc}
Consider the pinhole projection of the center of one gaussian whose position relative to the camera is $x,y,z$:
\begin{align}
	\label{eq:uv}
	(u, v) &= (\frac{f_x}{z} x, \frac{f_y}{z} y)
\end{align}
With $f_x,f_y$ the focal lengths of the camera, along $u,v$ axis.
We can see from eq. \ref{eq:uv} that any modification of the camera parameters that keeps $(f_x / z, f_y / z)$ constant will have no
impact of $(u,v)$. If all gaussians had the same $z$, it would not be possible to disentangle the parameters.
Though this case never arises, $f_x, f_y$ and $z$ are still highly correlated.
This correlation is related to the similarity between getting closer to the scene and zooming in with the camera. Note that in
3DGS \cite{kerbl20233d}, the focal $f_x, f_y$ parameters are replaced by the field of view $\phi_x,\phi_y$.

To help training the model, the parameters $(z_c, \phi_x, \phi_y)$ of a camera are replaced by $(a, b, c)$.
One reparameterization is to choose $(a, b, c)$ as the coordinates in the referential defined by the eigen vectors of
the Hessian matrix $\mathcal{H}$ of $\mathcal{L}(z_c,\phi_x,\phi_y)$. Calling $\mathcal{E}$ this matrix, we have:
\begin{align}
	\label{eq:abc}
	\begin{bmatrix}z_c\\\phi_x\\\phi_y\end{bmatrix} = \begin{bmatrix}z_{c}^0\\\phi_x^0\\\phi_y^0\end{bmatrix} + \begin{bmatrix}\mathcal{E}\end{bmatrix} \begin{bmatrix}a\\b\\c\end{bmatrix}
\end{align}

The rational behind this reparameterization is that in the neighborhood of the local minimum, the gradient of the
loss is approximated by $\frac{\partial\mathcal{L}}{\partial\theta} \approx \mathcal{H}\theta$.
By definition, in referential given by the Hessian eigen vectors, the Hessian matrix is diagonal, thus the derivative with respect to
one parameter is independent from the other parameters: $\frac{\partial\mathcal{L}}{\partial\theta_i} \approx \lambda_i\theta_i$
(with $\lambda_i$ the $i - th$ eigen value).
The Adam optimizer \cite{kingma2014adam} uses momentum, which is a way to accumulate 
a velocity vector in directions of persistent loss reduction
\cite{sutskever2013importance}. In the directions of low curvature, the loss gradient
tends to persist across iterations and be amplified by momentum accumulation. This enables efficient convergence
for parameters of low impact (small eigen values of the Hessian), that would otherwise be highly disturbed by high
impact parameters.

In practice, the first optim phase is done with the standard $(z_c,\phi_x,\phi_y)$ parameters, then the values of $(z_{c}^0,\phi_x^0,\phi_y^0)$ are set,
the hessian is evaluated, $(a,b,c)$ are initiated at $(0,0,0)$ and the rest of the optimization process is done with $(x_c,y_c,q,a,b,c)$ parameters.

Since the second derivation of the loss gives unstable results, we approximate the hessian by finite difference, evaluated from the gradient on the current
$(z_c,\phi_x,\phi_y)$ and on points around it $(z_c \pm \epsilon, \phi_x \pm \epsilon, \phi_y \pm \epsilon)$.

\section{Results}

\begin{figure}
	\includegraphics[width=8cm]{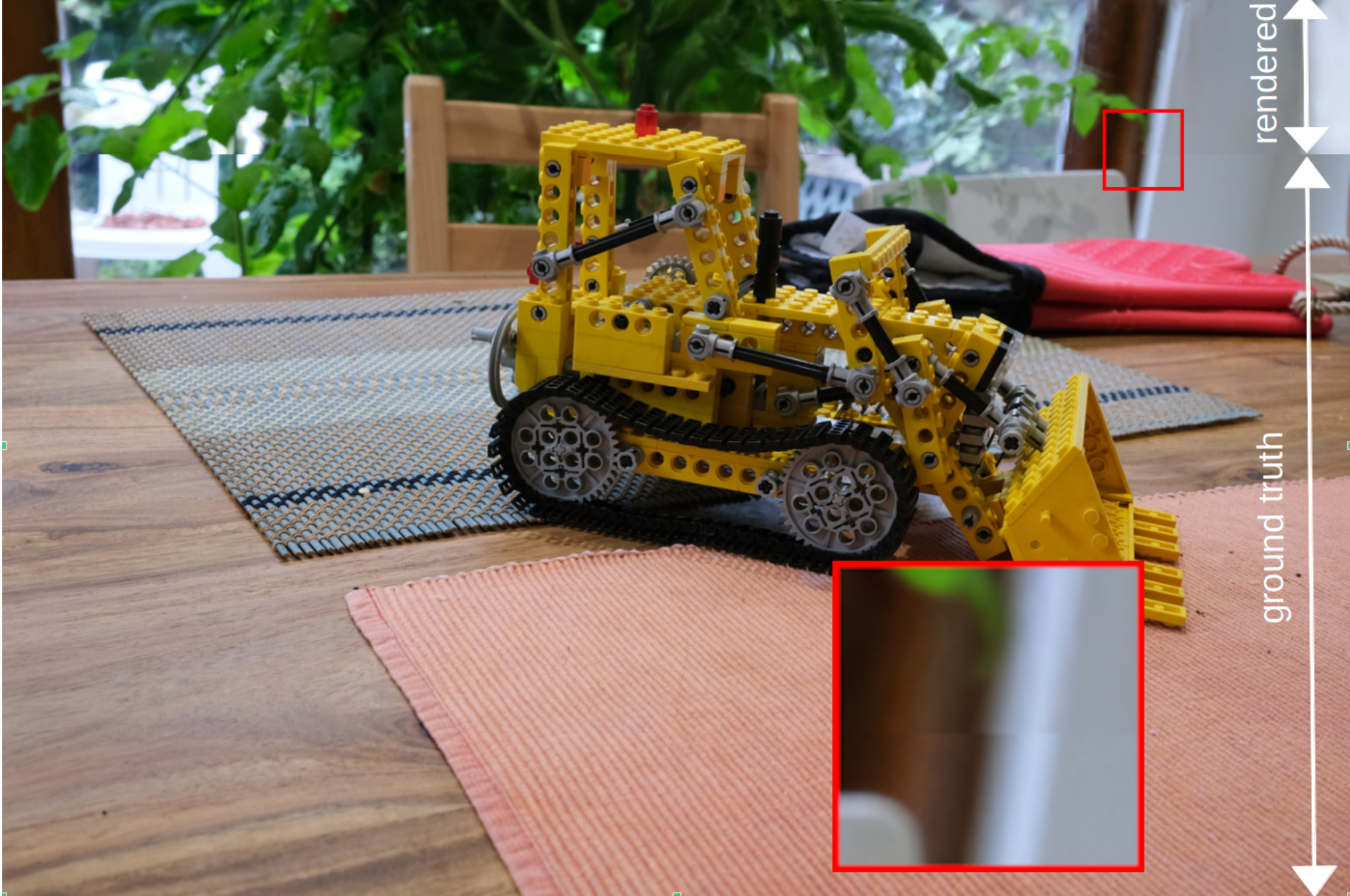}
  \caption{Example of failure. A detailed zone (the tree), seen by few cameras lead to bad reconstruction, which disturbs camera fine-tuning.}
  \label{img:quali4}
\end{figure}

Our experiments are processed in two steps: we first train the camera positions and model together, as described in \ref{method},
then we re-train the model, render the test images and process metrics with the original code of \cite{kerbl20233d}.
The results are listed in Table \ref{table:results}. 
Table \ref{table:n_vertices} gives the number of gaussians of the 3DGS \cite{kerbl20233d} for the same scenes.
The fine-tuned calibration enables generally a better PSNR with less gaussians.
The few images with worse PSNR are points of views shared by so few cameras that the reconstruction is not reliable. Cf Figure \ref{img:quali4}.

\begin{table}[h]
  \small
	\begin{center}
		{\rowcolors{2}{white}{lightgray}
		\begin{tabular}{|p{3cm}|c|c|}
			\hline
			scene & \shortstack{dataset calibration \\ PSNR [dB]} & \shortstack{ours \\ PSNR [dB]} \\
			\hline
			mip360 bicycle & 25.61 & \textbf{26.34} \\
			mip360 bonsai & 32.32 & \textbf{32.72} \\
			mip360 counter & 29.11 & \textbf{29.26} \\
			mip360 garden & 27.77 & \textbf{28.32} \\
			mip360 kitchen & \textbf{31.55} & 31.50 \\
			mip360 room & 31.72 & \textbf{31.94} \\
			mip360 stump & 26.90 & \textbf{27.28} \\
			T\&T Train & 22.12 & \textbf{22.61} \\
			T\&T Truck & 25.84 & \textbf{26.18} \\
			DB DrJohnson & 29.11 & \textbf{30.12} \\
			DB Playroom & 29.96 & \textbf{30.77} \\
			\hline
			Average & 28.36 & \textbf{28.79} \\
			\hline
			\end{tabular}
			}
			\caption{PSNR on the reference scenes as defined by 3DGS \cite{kerbl20233d}.
			Column "dataset calibration" lists results of training using the Colmap calibration provided with the datasets.
			Column "ours" lists results of training with the exact same code but our calibration.}
			\label{table:results}
		\end{center}
	\end{table}
	
  \begin{table}[h]
    \small
    \begin{center}
      {\rowcolors{2}{white}{lightgray}
      \begin{tabular}{|p{2.5cm}|c|c|}
        \hline
        scene & \shortstack{dataset calibration \\ gaussians [M]} & \shortstack{ours \\ gaussians [M]} \\
        \hline
        mip360 bicycle & 5.79 & \textbf{4.72} \\
        mip360 bonsai & 1.25 & \textbf{1.14} \\
        mip360 counter & 1.17 & \textbf{1.13} \\
        mip360 garden & 5.09 & \textbf{3.63} \\
        mip360 kitchen & 1.74 & \textbf{1.54} \\
        mip360 room & 1.49 & \textbf{1.38} \\
        mip360 stump & 4.66 & \textbf{4.26} \\
        T\&T Train & \textbf{1.01} & 1.13 \\
        T\&T Truck & 2.31 & \textbf{2.03} \\
        DB DrJohnson & \textbf{3.31} & 3.34 \\
        DB Playroom & 2.31 & \textbf{1.96} \\
        \hline
        Average & 2.35 & \textbf{2.06} \\
        \hline
        \end{tabular}
        }
        \caption{Number of gaussians (in millions) on the reference scenes as defined by 3DGS \cite{kerbl20233d}.
        Column "dataset calibration" lists results of training using the Colmap calibration provided with the datasets.
        Column "ours" lists results of training with the exact same code but our calibration.}
        \label{table:n_vertices}
      \end{center}
    \end{table}
    
\begin{figure}[h]
	\includegraphics[width=8cm]{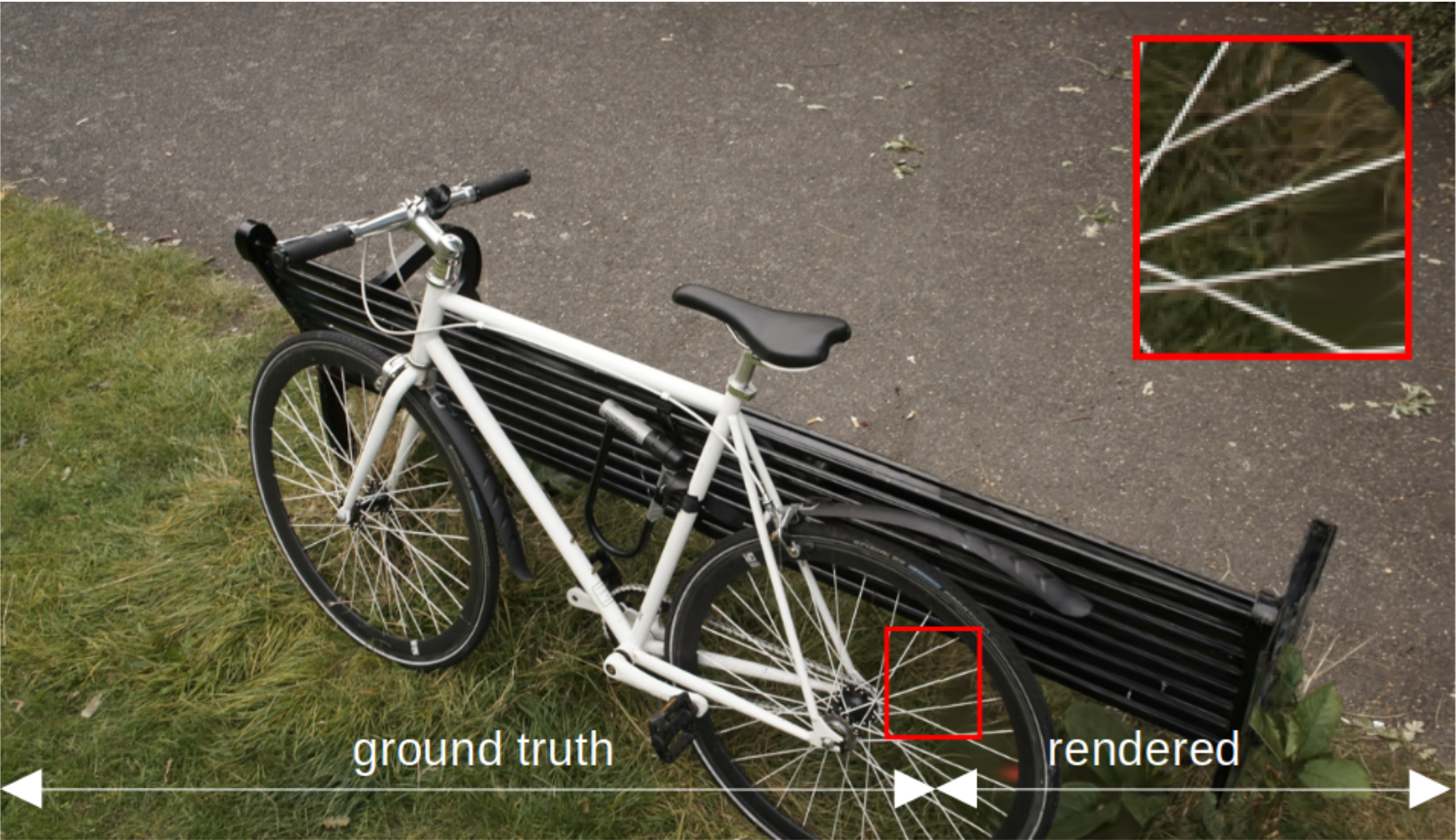}
	\includegraphics[width=8cm]{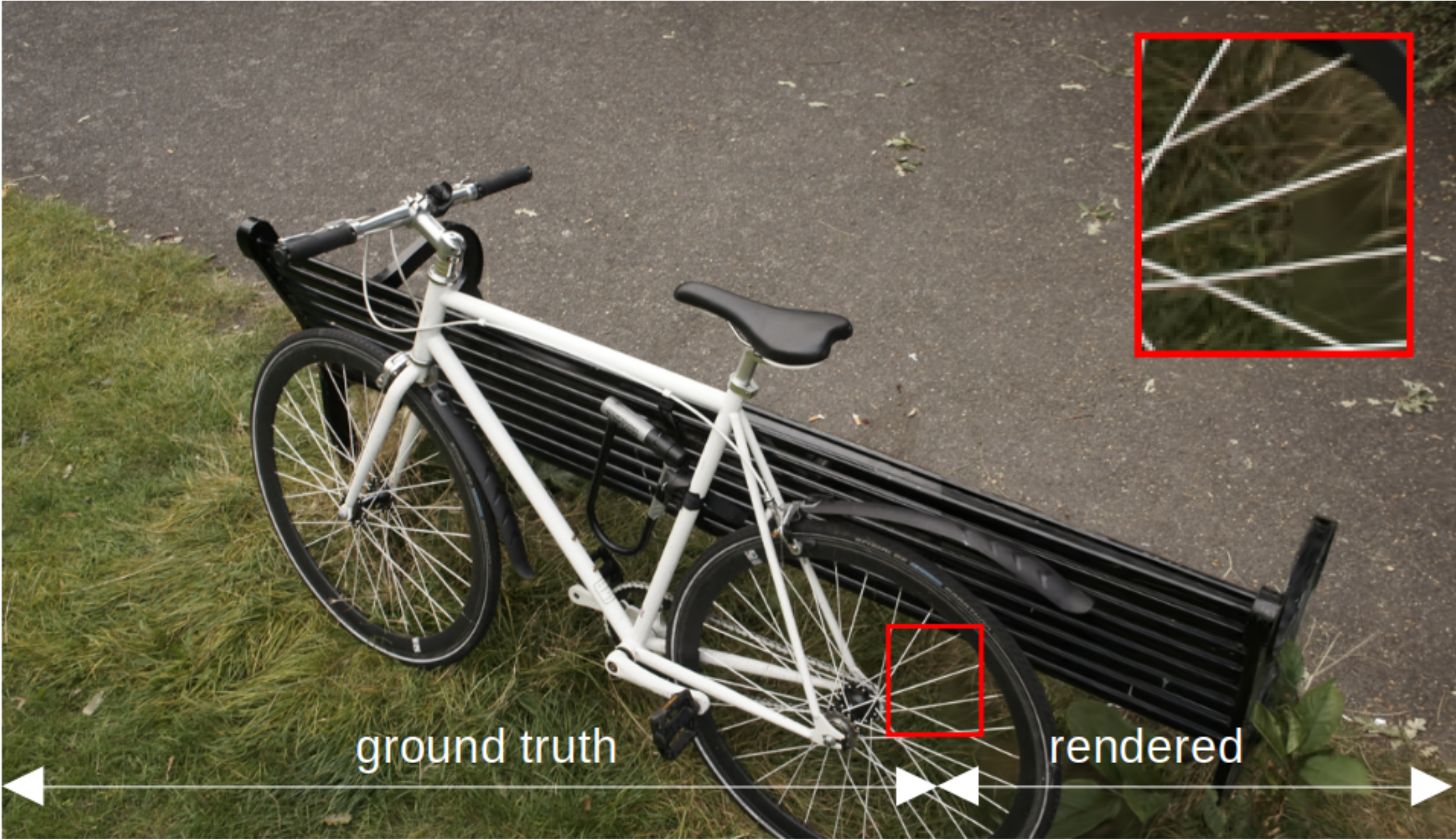}
	\caption{Reference (top) vs ours (bottom) on bicycle view.}
	\label{img:quali1}
\end{figure}

\begin{figure}[h]
  \includegraphics[width=8cm,height=4.6cm]{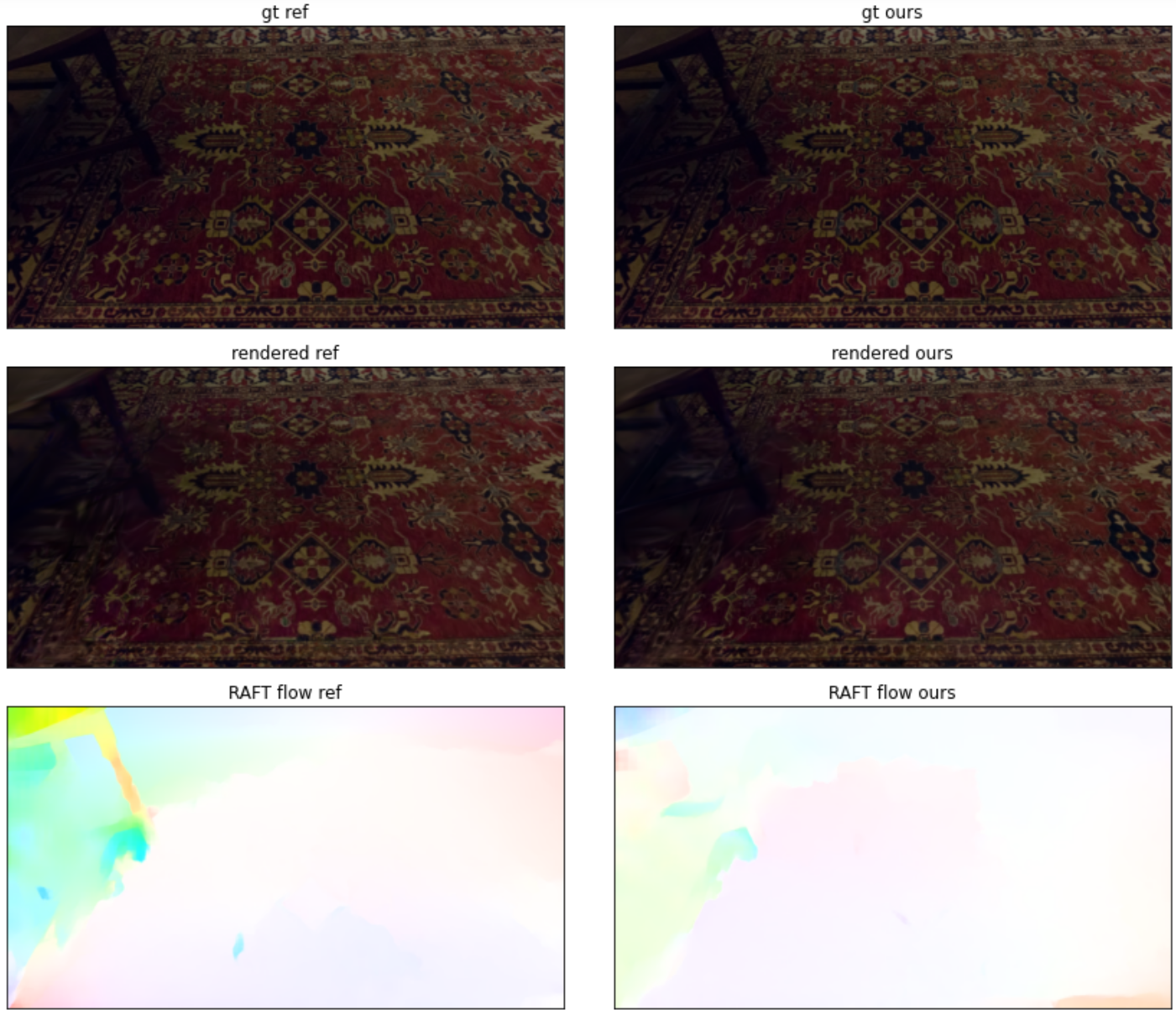}
\caption{Reference (left) vs ours (right) on DrJohnson view. Bottom: motion flow estimated with RAFT \cite{DBLP:journals/corr/abs-2003-12039}}
\label{img:quali2}
\end{figure}

\begin{figure}
  \includegraphics[width=7cm,height=3.4cm]{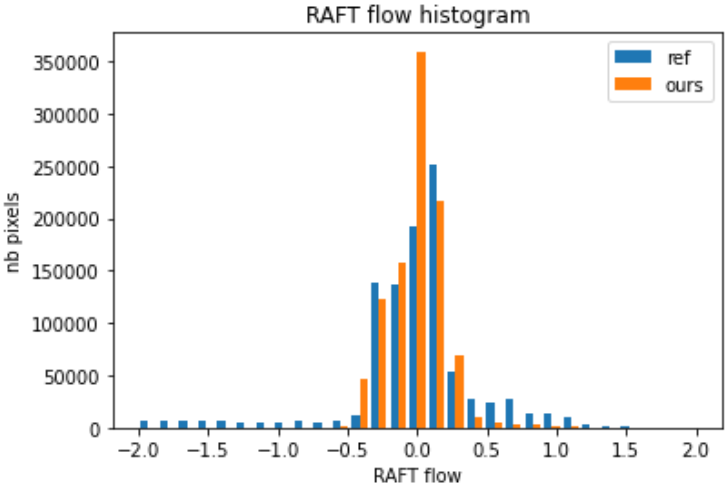}
\caption{Reference (blue) vs ours (orange) motion flow histogram on DrJohnson view. Motion is in pixels.}
\label{img:quali3}
\end{figure}

  On some images, one can see the slight calibration error of the Colmap input, and the result of the fine-tuning.
	Figure \ref{img:quali1} shows a slight misalignment on Colmap \cite{schonberger2018robust}, visible on the wheel
	of the bicycle \cite{barron2022mip}, which is corrected by camera fine-tuning.
  Figure \ref{img:quali2} shows another example on DrJohnson \cite{HPPFDB18} where we have used RAFT \cite{DBLP:journals/corr/abs-2003-12039}
  motion flow to help visualize the improvement between the calibration given with the datasets (ref) and the fine-tuned
  calibration (ours). Figure \ref{img:quali3} displays the motion histogram, in pixels.
	
\subsection{Ablations}

Table \ref{table:ablation} gives the PSNR obtained by the same process as ours, but with one of the aspects missing.
\vbox{%
\begin{itemize}
	\item w/o abc does not use reparameterization.
	\item w/o fov does not train ($\phi_x, \phi_y$). Note that this implies not to use reparameterization neither.
	\item L1 uses L1 loss instead of L2, for camera parameters training.
	\item once runs only on phase of cameras fine-tuning, instead of one phase every 3000 3DGS steps
\end{itemize}}

\begin{table}[h]
  \small
	\begin{center}
    {\rowcolors{2}{white}{lightgray}
    \begin{tabular}{|p{1.2cm}|c|c|c|c|c|}
      \hline
				scene  & w/o abc & w/o fov & L1 & once & full \\
				\hline
				bicycle & 25.81 & 25.74 & 25.82 & 25.62 & \textbf{26.34} \\
				bonsai & 32.31 & 32.15 & 32.41 & 32.32 & \textbf{32.72} \\
				counter & 29.04 & 29.09 & 29.13 & 29.25 & \textbf{29.26} \\
				garden & 27.70 & 27.50 & 27.72 & 27.87 & \textbf{28.32} \\
				kitchen & 31.12 & 31.07 & 30.71 & 31.61 & \textbf{31.14} \\
				room & 31.75 & 31.77 & 31.80 & \textbf{32.03} & 31.94 \\
				stump & 26.72 & 26.60 & 26.73 & 26.88 & \textbf{27.28} \\
				Train & 22.47 & 22.31 & 22.54 & 22.46 & \textbf{22.61} \\
				Truck & 26.13 & 26.05 & 26.11 & 25.86 & \textbf{26.18} \\
				DrJohnson & 29.93 & 29.55 & 29.81 & 29.47 & \textbf{30.12} \\
				Playroom & 30.61 & 30.36 & 30.45 & 29.99 & \textbf{30.77} \\
        \hline
        Average & 28.51 & 28.38 & 28.48 & 28.49 & \textbf{28.79} \\
        \hline
			\end{tabular}
  }
	\caption{Ablation: PSNR [dB] with some of the solution missing. Scenes bicycle, bonsai, counter, garden, kitchen, room and stump
  from Mip-nerf 360 \cite{barron2022mip}, Train and Truck from Tanks and temples \cite{knapitsch2017tanks}, DrJohnson and
  Playroom from Deep Blending \cite{HPPFDB18}.}
	\label{table:ablation}
	\end{center}
\end{table}

\section{Conclusion}

Considering the precision of camera calibration as a matter of importance for the comparison of publications of
novel view synthesis, we propose a solution to fine tune this calibration. The camera calibration uses a 
3DGS model which is trained in parallel in an interleaved training schedule.
The new calibration alone brings an average improvement of 0.4 dB PSNR on the dataset used as reference by 3DGS \cite{kerbl20233d}.
The fine tuning may be long and its suitability depends on the criticity of training time, but for calibration
of reference scenes, such as Mip-nerf 360 \cite{barron2022mip}, the stake of novel view quality is the most important.

\clearpage

\bibliographystyle{plain}
\bibliography{calibration}

\begin{thebibliography}{10}

\bibitem{barron2022mip}
Jonathan~T Barron, Ben Mildenhall, Dor Verbin, Pratul~P Srinivasan, and Peter Hedman.
\newblock Mip-nerf 360: Unbounded anti-aliased neural radiance fields.
\newblock In {\em Proceedings of the IEEE/CVF conference on computer vision and pattern recognition}, pages 5470--5479, 2022.

\bibitem{HPPFDB18}
Peter Hedman, Julien Philip, True Price, Jan-Michael Frahm, George Drettakis, and Gabriel Brostow.
\newblock Deep blending for free-viewpoint image-based rendering.
\newblock {\em ACM Transactions on Graphics (SIGGRAPH Asia Conference Proceedings)}, 37(6), November 2018.

\bibitem{kerbl20233d}
Bernhard Kerbl, Georgios Kopanas, Thomas Leimk{\"u}hler, and George Drettakis.
\newblock 3d gaussian splatting for real-time radiance field rendering.
\newblock {\em ACM Trans. Graph.}, 42(4):139--1, 2023.

\bibitem{kheradmand20243d}
Shakiba Kheradmand, Daniel Rebain, Gopal Sharma, Weiwei Sun, Jeff Tseng, Hossam Isack, Abhishek Kar, Andrea Tagliasacchi, and Kwang~Moo Yi.
\newblock 3d gaussian splatting as markov chain monte carlo.
\newblock {\em arXiv preprint arXiv:2404.09591}, 2024.

\bibitem{kingma2014adam}
Diederik~P Kingma and Jimmy Ba.
\newblock Adam: A method for stochastic optimization.
\newblock {\em arXiv preprint arXiv:1412.6980}, 2014.

\bibitem{knapitsch2017tanks}
Arno Knapitsch, Jaesik Park, Qian-Yi Zhou, and Vladlen Koltun.
\newblock Tanks and temples: Benchmarking large-scale scene reconstruction.
\newblock {\em ACM Transactions on Graphics (ToG)}, 36(4):1--13, 2017.

\bibitem{lindenberger2021pixel}
Philipp Lindenberger, Paul-Edouard Sarlin, Viktor Larsson, and Marc Pollefeys.
\newblock Pixel-perfect structure-from-motion with featuremetric refinement.
\newblock In {\em Proceedings of the IEEE/CVF international conference on computer vision}, pages 5987--5997, 2021.

\bibitem{schonberger2018robust}
Johannes~L Sch{\"o}nberger.
\newblock {\em Robust methods for accurate and efficient 3D modeling from unstructured imagery}.
\newblock PhD thesis, ETH Zurich, 2018.

\bibitem{sutskever2013importance}
Ilya Sutskever, James Martens, George Dahl, and Geoffrey Hinton.
\newblock On the importance of initialization and momentum in deep learning.
\newblock In {\em International conference on machine learning}, pages 1139--1147. pmlr, 2013.

\bibitem{DBLP:journals/corr/abs-2003-12039}
Zachary Teed and Jia Deng.
\newblock {RAFT:} recurrent all-pairs field transforms for optical flow.
\newblock {\em CoRR}, abs/2003.12039, 2020.

\bibitem{wang2024dust3r}
Shuzhe Wang, Vincent Leroy, Yohann Cabon, Boris Chidlovskii, and Jerome Revaud.
\newblock Dust3r: Geometric 3d vision made easy.
\newblock In {\em Proceedings of the IEEE/CVF Conference on Computer Vision and Pattern Recognition}, pages 20697--20709, 2024.

\bibitem{ye2024gsplat}
Vickie Ye, Ruilong Li, Justin Kerr, Matias Turkulainen, Brent Yi, Zhuoyang Pan, Otto Seiskari, Jianbo Ye, Jeffrey Hu, Matthew Tancik, et~al.
\newblock gsplat: An open-source library for gaussian splatting.
\newblock {\em arXiv preprint arXiv:2409.06765}, 2024.

\bibitem{zeller2024gsplatloc}
Atticus~J Zeller.
\newblock Gsplatloc: Ultra-precise camera localization via 3d gaussian splatting.
\newblock {\em arXiv preprint arXiv:2412.20056}, 2024.

\end{thebibliography}

\end{document}